\documentclass[sigconf]{acmart}
\DeclareUnicodeCharacter{0394}{$\Delta$}

\AtBeginDocument{%
  }
\renewcommand\footnotetextcopyrightpermission[1]{} 
\usepackage{multirow}
\usepackage{xspace}

\usepackage{booktabs}
\usepackage{bm}

\usepackage{amssymb,amsthm,amsfonts}
\usepackage{enumerate,enumitem}
\usepackage{diagbox}
\usepackage{booktabs}
\usepackage{graphicx}
\usepackage{xcolor}
\usepackage{amsmath}
\usepackage{array}
\usepackage{caption}
\usepackage[most]{tcolorbox}
\usepackage{xcolor}
\usepackage{threeparttable}
\usepackage{colortbl}

\setcopyright{none}
\makeatletter
\renewcommand\@formatdoi[1]{\ignorespaces}
\makeatother
\newcommand{\sys}{{\texttt{HolmesEye}}\xspace}
\newcommand{\data}{{\texttt{PAPI}}\xspace}

\newenvironment{icompact}{
  \begin{list}{$\bullet$}{
    \itemindent 0em
    \itemsep 0pt
    \leftmargin 0.15in}
      }
{\normalsize
\end{list}
}

\begin{document}

\title{The Eye of Sherlock Holmes: Uncovering User Private Attribute Profiling via Vision-Language Model Agentic Framework}
% \title{The Eye of Sherlock Holmes: Uncovering User Private Attribute Profiling via Multimodal LLM Agentic Framework}

%%

\author{Feiran Liu$^1$, Yuzhe Zhang$^2$, Xinyi Huang$^2$, Yinan Peng$^3$, Xinfeng Li$^{1,\dag}$, Lixu Wang$^1$,\\ Yutong Shen$^2$, Ranjie Duan$^4$, Simeng Qin$^1$, Xiaojun Jia$^1$, Qingsong Wen$^5$, Wei Dong$^1$
}

\affiliation{
$^1$Nanyang Technological University, $^2$Beijing University of Technology,  $^3$Hengxin Tech, $^4$Alibaba Group, $^5$Squirrel Ai Learning
  \\ \country{}
}

\renewcommand{\shortauthors}{Feiran Liu et al.}
\newcommand\blfootnote[1]{%
  \begingroup
  \renewcommand\thefootnote{}\footnote{#1}%
  \addtocounter{footnote}{-1}%
  \endgroup
}



%%%%%%%%% ABSTRACT
\begin{abstract}
Our research reveals a new privacy risk associated with the vision-language model (VLM) agentic framework: the ability to infer sensitive attributes (e.g., age and health information) and even abstract ones (e.g., personality and social traits) from a set of personal images, which we term ``image private attribute profiling.'' 
This threat is particularly severe given that modern apps can easily access users’ photo albums, and inference from image sets enables models to exploit inter-image relations for more sophisticated profiling. 
However, two main challenges hinder our understanding of how well VLMs can profile an individual from a few personal photos: (1) the lack of benchmark datasets with multi-image annotations for private attributes, 
and (2) the limited ability of current multimodal large language models (MLLMs) to infer abstract attributes from large image collections. 
In this work, we construct \data, the largest dataset for studying private attribute profiling in personal images, comprising 2,510 images from 251 individuals with 3,012 annotated privacy attributes.
We also propose \sys, a hybrid agentic framework that combines VLMs and LLMs to enhance privacy inference. 
\sys uses VLMs to extract both intra-image and inter-image information and LLMs to guide the inference process as well as consolidate the results through forensic analysis, overcoming existing limitations in long-context visual reasoning. 
Experiments reveal that \sys achieves a 10.8\% improvement in average accuracy over state-of-the-art baselines and surpasses human-level performance by 15.0\% in predicting abstract attributes.
This work highlights the urgency of addressing privacy risks in image-based profiling and offers both a new dataset and an advanced framework to guide future research in this area.
\blfootnote{$^{*}$Xinfeng Li is the corresponding authors. Email: lxfmakeit@gmail.com}
\end{abstract}

\maketitle

%%%%%%%%% BODY TEXT
\section{Introduction}

The rise of multimodal large language models (MLLMs) has transformed artificial intelligence, with vision-language models (VLMs) now capable of sophisticated reasoning across visual and textual domains \cite{minaee2024large, xu2021vlm, liuyue_efficient_reasoning,lu2025damo}. These advances have produced remarkable capabilities from answering complex visual questions to generating detailed image analyses---yet they also introduce new privacy concerns \cite{delgado2022survey, staab2023beyond, wang2025safety, wang2025comprehensive,li2025tuni}. This evolution raises a critical question: \textit{What sensitive information can be extracted from seemingly ordinary images when analyzed by increasingly capable AI systems?}

Our research reveals a concerning answer to this question: modern VLMs, when strategically employed, can construct detailed personal profiles from collections of ordinary images, inferring not only concrete attributes (e.g., age, sex, and location) \cite{yan2020mitigating}, but also abstract ones (e.g., personality traits, socioeconomic status, and behavioral patterns). This capability---what we term \textit{image private attribute profiling}---represents a fundamental shift in privacy risk assessment. Unlike traditional identification threats \cite{meloy2015concept}, attribute profiling leverages subtle patterns and contextual relationships across multiple images to extract sensitive information that users have never explicitly shared or consented to reveal \cite{jayaraman2022attribute}.

The privacy implications of this capability are particularly concerning in today's digital landscape. Over 95\% of mobile applications request access to photos, and more than 3.2 billion images are shared daily across social media platforms \cite{JOZANI2020106260}. As users continuously generate vast amounts of personal visual data, each shared image contributes to a larger, often unintended, disclosure of personal information. This creates an unprecedented privacy challenge: even if users are careful to avoid sharing sensitive details in individual photos, the overall patterns across their image collections may still reveal deeply personal attributes \cite{zhang2022attribute}.

\begin{figure}[t]
\centering
\includegraphics[width=\columnwidth]{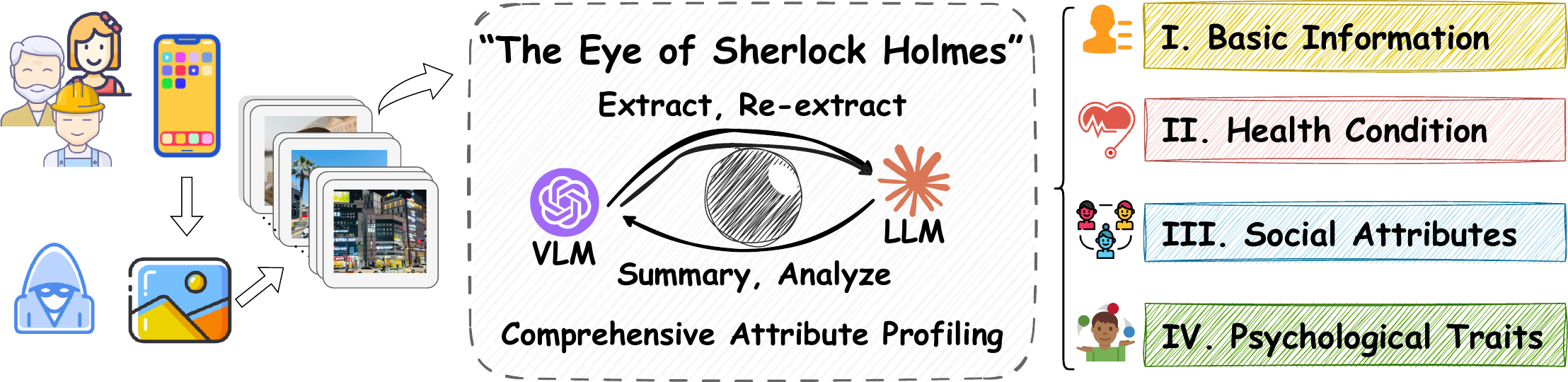}
\vspace{-10pt}
\caption{\sys develops VLM perception and LLM reasoning to extract private attributes from image collections, analyzing both individual images and cross-image patterns to profile users across four attribute domains.}
\vspace{-10pt}
\label{fig:pipeline}
\end{figure}

This emerging privacy threat has remained largely unexplored due to two significant research challenges. First, there is a lack of benchmark datasets to study how profiles can be constructed from multiple images. Existing visual datasets typically focus on single-image attribute recognition \cite{wang2017attribute, abraham2019fairness} with limited privacy dimensions, which are insufficient for investigating complex cross-image inference attacks. Second, existing approaches primarily focus on analyzing individual images in isolation, lacking the tools to explore inter-image relationships \cite{goossen2009medical} and contextual reasoning that enable sophisticated profiling. These challenges have hindered our understanding of privacy vulnerabilities and the development of effective protective measures.

To address these critical gaps, we develop \data (Private Attributes in Personal Images), a benchmark dataset containing 2,510 images from 251 individuals with 3,012 (251$\times$12) person-attribute annotations. \data spans 12 diverse privacy attributes across four domains: basic information, health condition, social attributes, and psychological traits. \data serves as a standard for constructing multi-image personal profiles \cite{kim2023privacy}. Its data collection complies with GDPR (General Data Protection Regulation) requirements through anonymization techniques, consent-based collection, and privacy-preserving processing, ensuring ethical research practices while enabling meaningful investigations into privacy risks.

Building on this dataset, we propose \sys, a hybrid agentic framework \cite{zheng2017consensus} inspired by the deductive reasoning methods of Sherlock Holmes (see Figure \ref{fig:pipeline}). \sys combines the strengths of VLMs and LLMs in a coordinated process. VLMs extract both details from individual images (i.e., intra-image) and relationships across multiple images (i.e., inter-image) \cite{zhang2021dmrfnet}, while LLMs guide the inference process through forensic analysis \cite{jin2022towards}. Together, these agents create detailed profiles of individuals, particularly good at inferring abstract attributes like personality (MBTI) and leadership tendency \cite{stapel1996referents}, that require reasoning across multiple contexts.

Extensive experiments demonstrate that \sys significantly outperforms baseline methods across all attribute categories, achieving a 10.8\% improvement in average accuracy. Moreover, \sys surpasses human-level performance by 15.0\% in predicting abstract attributes. Notably, while human analysis takes an average of 11.5 minutes to profile an individual, \sys accomplishes the same task in just about 3.5 minutes, suggesting how modern MLLMs \cite{wang2024exploring} can be weaponized for efficient, large-scale privacy intrusions. We also discuss \sys's limitation and potential countermeasures to mitigate the privacy risks posed by vision-language agents.

The key contributions of our work are:

\begin{icompact}
    \item We identify a new privacy threat: the detailed profiling of personal attributes from collections of everyday images, showing how normal visual data can reveal private information.
    
    \item We construct \data, a GDPR-compliant benchmark dataset for studying multi-image private attribute profiling, providing a foundation for research on privacy risks.
    
    \item We develop \sys, an agentic framework for complex, cross-modal reasoning of private attributes, which integrates VLMs and LLMs within a structured, forensic workflow.
    
    \item We perform thorough experiments, including ablation studies, impact analysis, and comparisons to baseline models and human analysis, confirming \sys's superior performance in attribute profiling.
\end{icompact}
\section{Background and Relative Works}
Our research builds upon work across two interconnected domains that provide essential context for understanding privacy implications of advanced visual intelligence systems.

\subsection{Attribute Inference Attack}
% 1. definition
% 2. related works
% 3. Significance of our work.
Attribute inference attack (AIA) aims to deduce sensitive attribute values of an individual by analyzing accessible data, which may not directly disclose such private information \cite{hu2022membership, zhao2021feasibility, wu2024inference}. AI systems, such as Vision-Language Models (VLMs), may infer private facts from photographs by analyzing subtle indications such as facial expressions, attire, and background features \cite{tomekce2024private}. Traditionally, AIA has concentrated on structured data, such as text, where correlations between observed elements might disclose sensitive information \cite{zhang2022attribute}. However, advanced AI models can already infer such properties from visual data by detecting hidden patterns, even when they are not directly present in the image \cite{delgado2022survey}.

AIA research has grown with the development of large language models (LLMs), which have shown the capacity to infer private characteristics from text data by evaluating patterns and context \cite{jayaraman2022attribute, wu2024inference, staab2023beyond}. Early image research focused on explicit identification tasks like facial recognition and location detection \cite{cheng2022personal, li2023security, meloy2015concept}. Tömekçe et al. \cite{tomekce2024private} demonstrated that VLMs can extract sensitive information from photos, even if it is not explicitly presented. This work highlights a paradigm shift in privacy concerns by revealing the capability of modern AI systems to extract sensitive information from seemingly innocent photographs \cite{baruh2017online}.

Our study expands on this foundation by focusing on comprehensive and detailed user profiling. Rather than inferring an attribute, e.g., age, from a given image, we investigate how multiple images can be analyzed collectively to reveal both concrete and abstract sensitive characteristics of an individual, such as health condition or psychological traits \cite{stapel1996referents, ullah2020protecting}. To advance this field, we introduce \data, the first benchmark dataset for multi-image private attribute profiling, which includes thousands of annotated photos spanning a diverse range of privacy attributes \cite{abraham2019fairness, terwee2010qualitative}. Furthermore, we develop \sys, a MLLM agentic framework that combines VLMs and LLMs to enhance the inference process by analyzing both intra- and inter-image details \cite{goossen2009medical, ma2024spatialpin, bye2009design}. Our approach significantly improves the accuracy of private attributes inference and introduces a novel methodology to analyze the privacy risks posed by image collections \cite{kim2023privacy}.

\subsection{Vision-Language Models}

Vision-Language Models (VLMs) represent a significant advancement in multimodal AI, integrating visual perception with natural language understanding capabilities \cite{xu2021vlm, wang2024exploring}. These models process and reason about visual and textual information simultaneously, enabling sophisticated cross-modal understanding. Leading VLMs include ChatGPT \cite{openai2024gpt4statement}, Gemini \cite{gemini2024multimodal}, and Claude \cite{Anthropic2025}, which have demonstrated remarkable abilities in visual reasoning, image description, and multimodal analysis tasks \cite{minaee2024large, zhang2024mmllms}.

VLMs are typically deployed for a range of tasks including image captioning, visual question answering, and multimodal content generation \cite{zhang2024mmllms}. When applied to attribute recognition, these models offer significant advantages over traditional computer vision approaches due to their emergent reasoning capabilities~\cite{li2025system,fang2025safemlrm} to understand contextual information \cite{yan2020mitigating, wang2017attribute}. Unlike specialized models that focus solely on detecting visible attributes like age or clothing, VLMs can infer latent characteristics from contextual cues and environmental factors present in images \cite{zhao2024survey}. However, these models are not without limitations—they may generate incomplete or hallucinated descriptions of visual content, miss fine-grained details crucial for accurate attribute inference, and struggle with long-context reasoning across multiple images \cite{li2023loogle}.

We reveal a new privacy risk posed by VLMs when strategically employed for attribute profiling from ordinary image collections \cite{tomekce2024private, ullah2020protecting}. Specifically, \sys synergistically combines VLMs' visual perception capabilities with LLMs' reasoning abilities \cite{zheng2017consensus, thorisson2007integrated, zhang2021dmrfnet}. This agentic framework enables more accurate inference of personal attributes from multiple images than VLM-only attack could achieve. Third, we empirically demonstrate the severity of this privacy threat by showing that our system can construct detailed personal profiles with higher accuracy than human analysis while requiring significantly less time \cite{wilson2015evaluation}, highlighting the urgent need for new privacy protection mechanisms \cite{baruh2017online, kim2023privacy}.

\section{Threat Model}\label{sec:threat_model}
\noindent \textbf{Attack Scenarios.} With the widespread use of social media, users often post their photos, making them increasingly vulnerable to visual-driven private attribute analysis \cite{delgado2022survey}. Several apps exploit various methods to steal users' photo privacy and conduct complex privacy analysis \cite{ye2020exploiting, zuo2019does}. Common scenarios where attackers may collect sensitive photo data include:
\begin{icompact}
\item \textbf{Social Media Data Collection:} Attackers scrape image data from social media platforms to gather sensitive information \cite{wang2019hierarchical}.
\item \textbf{Cloud Storage Vulnerabilities:} Users may store their personal photos in cloud services, which, if compromised, allow unauthorized access to large volumes of private images \cite{zuo2019does}.
\item \textbf{Illegal Access to Albums:} Some smartphone apps unauthorizedly access users' photo albums and collect sensitive data~\cite{ye2020exploiting}.
\end{icompact}

\noindent \textbf{Adversary's Goal.}
The attacker's objective is to infer private attributes from a set of personal images as accurately and comprehensively as possible \cite{jayaraman2022attribute, wu2024inference}. While some attributes, such as sex, may be directly visible, the primary focus is on extracting indirect private information, such as Psychological Traits and Social Attributes, which are embedded in image collections but not explicitly disclosed \cite{zhang2022attribute}. The attacker aims to build a detailed profile by analyzing both individual images and the relationships between multiple images, going beyond explicit details shared in any single image. We summarize the primary targets of the attacker as the following four-group victim attributes. This profile can then be monetized through direct financial profit (such as selling to data brokers) or used to design targeted phishing attacks, sophisticated social engineering, fraud, or other criminal activities~\cite{ullah2020protecting}.

\begin{tcolorbox}[colframe=black!25, colback=gray!10, coltitle=black, title=Private Attributes of Adversaries' Interests, center title]
\label{box:privateAttributes}
\textbf{I. Personal Basic Information:} 
Age (AG), Sex (SE), Region (RE) \cite{carvajal2013associations}\\
\textbf{II. Health Condition:} 
Hygiene Habits (HH), Daily Routines (DR), Health Status (HS) \cite{salvador2013basic}\\
\textbf{III. Social Attributes:}
Consumption Ability (CA), Income Level (IN), Occupation (OC) \cite{baruh2017online}\\
\textbf{IV. Psychological Traits:}
MBTI (MBTI), Social Activity Level (SA), Leadership Tendency (LT) \cite{stapel1996referents}
\end{tcolorbox}

\noindent \textbf{Adversary's Knowledge and Capability.} The focus of this study is to understand how existing VLMs and LLMs can be used to perform privacy inference attacks on personal image collections \cite{wang2024exploring, tomekce2024private}. We assume the attacker has access to these MLLMs---whether open-source or proprietary~\cite{zhao2003quality}---and possesses the expertise to use them effectively. This includes designing prompts, switching models, and developing appropriate scheduling strategies for VLMs and LLMs for inferring private attributes from multiple images \cite{kong2023better, sun2025strong}. Additionally, the attacker may have image analysis skills, allowing them to preprocess image data using various tools to enhance the extraction of privacy-sensitive features \cite{thuraisingham1999technologies}, as well as expertise in private attribute analysis and forensic techniques \cite{zhang2022attribute, swiderski2004threat}.

\section{\data Dataset Construction}
\begin{figure*}[t!]
  \centering
  \includegraphics[width=\linewidth]{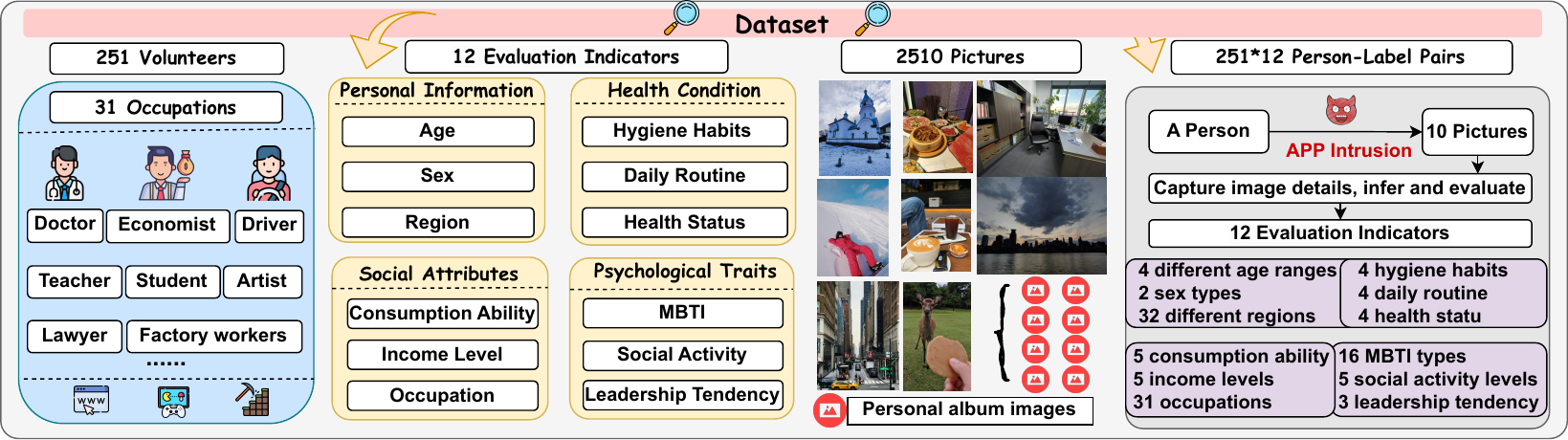}
  \vspace{-10pt}
  \caption{The PAPI dataset consists of 2,510 images collected from 251 volunteers across 31 occupations, with each individual represented by 10 photos. It includes 12 evaluation indicators across four categories: personal information, health condition, social attributes, and psychological traits.}
  \label{fig:dataset}
  \vspace{-10pt}
\end{figure*}

Existing datasets contain only limited annotations of sensitive attributes and are inadequate for analyzing private attributes across multiple images from the same individual~\cite{tomekce2024private}. To address this critical gap, we present \data, the first benchmark dataset specifically designed for multi-image privacy attribute profiling, comprising 2,510 images with 3,012 person-attribute annotations spanning four privacy domains.

\noindent \textbf{Key Features.} \data features 251 individuals, each represented by a carefully curated set of 10 images that indirectly reveal personal attributes. These attributes, as categorized in Section~\ref{sec:threat_model}~\cite{swiderski2004threat}, are not explicitly displayed but can be inferred through comprehensive analysis of both individual image content and cross-image contextual relationships~\cite{tomekce2024private, zhang2022attribute}.
\begin{icompact}
\item \textbf{Diverse Image Contexts.} To maximize ecological validity, PAPI incorporates images from a wide spectrum of real-life scenarios including professional environments, social gatherings, recreational activities, and everyday routines across various geographical and cultural settings~\cite{gdalyahu2001self}.
\item \textbf{Privacy Protection.} All data collection adheres to strict ethical guidelines with IRB approval. The dataset balances research utility with privacy preservation through appropriate anonymization techniques~\cite{ullah2020protecting}.
\end{icompact}

\noindent \textbf{Dataset Construction Process.} 
We systematically identified 31 representative occupations and recruited qualified volunteers across these professions~\cite{wilson2015evaluation}. Participants were instructed to select 10 photographs from their personal collections that subtly rather than explicitly reflected their attributes. They then provided ground-truth labels for 12 evaluation indicators based on their actual personal circumstances~\cite{abraham2019fairness, salvador2013basic}. To ensure data quality, our research team conducted verification of the image-attribute relationships, and any inconsistencies were resolved through follow-up discussions with participants\footnote{This work's research purpose and dataset collection methodology received approval from our institutional review board.}.
For enhanced privacy protection, some participants opted to provide similar publicly available images rather than their personal photographs. This approach ensures both privacy safeguards and dataset authenticity~\cite{baruh2017online, kim2023privacy}.
Detailed characteristics of the \data dataset are illustrated in Figure~\ref{fig:dataset}.
\section{Methodology}
\begin{figure*}[h]
  \centering
  \includegraphics[width=\linewidth]{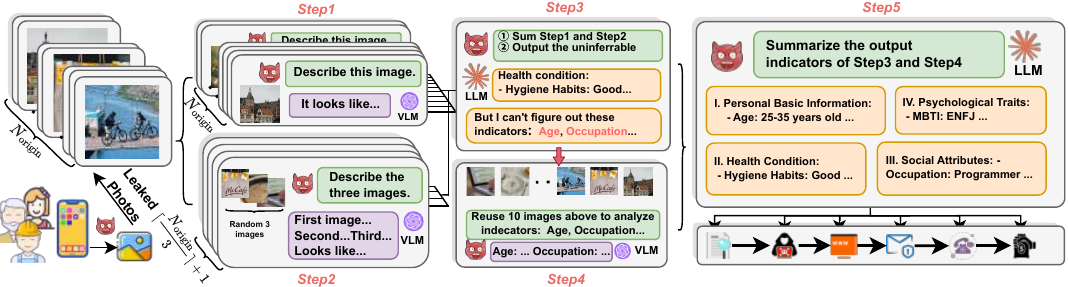}
  \vspace{-10pt}
  \caption{The process of inferring private information using the HolmesEye framework. Step 1 extracts individual image information. Step 2 analyzes inter-image relationships. Step 3 summarizes findings and identifies uninferable attributes. Step 4 inquiry missing indicators from all images. Step 5 consolidates results into comprehensive attribute profiles.}
  \label{fig:our_method}
  \vspace{-10pt}
\end{figure*}

\textbf{Overview.} To assess the risks associated with image privacy attribute analysis, we propose a framework called $\text{\sys}$, which includes multiple VLM and LLM agents that operate under the processes of extraction, analysis, summarization, inquiry, and decision-making, as shown in Figure~\ref{fig:our_method}. Specifically, VLM agent represented by $\phi_V$ and LLM agent represented by $\phi_L$. 

First, $\text{\sys}$ extracts preliminary information $\mathbf{F}_{\text{intra}}$
from a single image using the VLM agent. Next, the images are grouped, and specific prompts are used to analyze inter-image information $\mathbf{F}_{\text{inter}}$ from each group. After obtaining the initial reasoning results $\mathbf{F}_{\text{intra}}$ and $\mathbf{F}_{\text{inter}}$ from the VLM agent, the LLM agent assembles these results to infer the privacy attributes $\hat{\mathbf{A}}^{(\text{init})}$. 

Subsequently, the LLM agent inquires the VLM about any privacy information that cannot yet be inferred from the summary. Through the VLM’s responses to these inquiries, represented by \(q\), the LLM agent makes a final judgment $\mathbf{F}_{\text{intra}}$ regarding the privacy attributes, based on all previous inquiry and response information. If the initial inference $\hat{\mathbf{A}}^{(\text{init})}$ accounts for all the privacy attributes, it is considered the final result $\mathbf{F}_{\text{intra}}$.

\subsection{Extraction}
Based on previous work on VLM benchmark tests, although VLM performs relatively well on image data, there is still a significant gap when it comes to tasks that require in-depth and complex reasoning ~\cite{SchulzeBuschoff2024}. To improve the zero-shot reasoning capabilities of VLMs in fine-grained privacy inference tasks, we designed a prompt engineering strategy, such as identifying color patterns, object presence, or scene composition (see supplementary materia A.1). This guidance is then incorporated into a fixed prompt template, allowing the VLM to focus on attribute-relevant visual features while avoiding distraction from irrelevant details. The same prompt structure is applied uniformly across all image inputs to ensure consistency.

Specifically, we input \data's privacy attributes into an LLM agent and ask how to extract more relevant information. For example, the guidance generated could be, ``You should analyze the color features, content, and composition of the image.''
These guidelines, combined with our custom-designed prompt (P), help the VLM agent extract more detailed information from the image. Note that the prompt remains the same for each image extraction(see supplementary material A.1).

Mathematically, this can be expressed as:
\begin{equation}
F_{\text{intra}} ={ \phi_V\left(p^{\text{ex}}_V \oplus F_i\right)}_{i=1}^{i=N_{\text{origin}}}
\end{equation}
where $F_{\text{intra}}$ represents the extracted information, $\phi_V$ is the VLM agent, \(p_\mathrm{V}^\mathrm{ex}\) is the VLM's extraction prompt template, $F_i$ is the figure from same person, $N_{\text{origin}}$ is the number of images input and $\oplus$ denotes concatenation of prompt and auxiliary information before feeding into the model.

\subsection{Analysis}\label{sec:analysis}
As mentioned in Section \ref{sec:extraction}, the VLM can use guidance prompts to better extract relevant information, helping it focus on the most pertinent aspects of a task. Therefore, for tasks focused on extracting information between images, we apply the same method. Specifically, we input \data’s privacy attributes into an LLM agent, asking how to better focus on the relationships between objects in different images and the similarities and differences in their stylistic characteristics. The guidance might be: ``You should analyze the relationships between objects in different images and the similarities or differences in their styles.''

One major challenge in applying VLM to privacy attribute analysis is their limited ability to handle long-context responses. Even with VLM enhancements, the input of many images still does not allow for sufficient extraction. To address this, we divide the images set into groups of three images, allowing the VLM to effectively capture inter-image relationships while avoiding context overload. This three-image grouping strategy is empirically validated in our experiments, as shown in Section~\ref{exp:num} and Table~\ref{table:figure}, where it yields the highest attribute inference accuracy among tested group sizes.

Using this method, we can fully analyze the different images and extract a large amount of privacy information from the inter-image relationships. 

Mathematically:
\begin{equation}
F_{\text{inter}} = [\phi_V\left(p^{\text{analy}}_V \oplus \{F_i\}_{i=1}^{i=N_{\text{origin}}+1}\right)]_{j=1}^{j=4}
\end{equation}
where $F_{\text{inter}}$ represents the analyzed information, $\phi_V$ is the VLM agent, and $p_V^{\text{analy}}$ is the analysis prompt template.

\label{sec:extraction}

\subsection{Summarization}
\label{sec:summarization}
Currently, we have gathered a large amount of information from the images through the VLM agent. This information contains potential clues to infer our target privacy attributes, but since it has not been classified and summarized, it is currently disorganized. Some deeper information may not even have been inferred yet.

To address this, we leverage the LLM agent’s powerful reasoning ability to summarize all the extracted information. After classifying it according to privacy attributes, we can perform initial inferences, denoted as $\hat{A}^{(\text{init})}$. 

This is expressed as:
\begin{equation}
\hat{A}^{(\text{init})} = \phi_L\left(p^{\text{sum}}_L \oplus F_{\text{inter}}\oplus F_{\text{intra}}\right)
\end{equation}
where $\hat{A}^{(\text{init})}$ represents the summarized inference result, $\phi_L$ is the LLM agent, and $p_L^{\text{sum}}$ is the summarization prompt template.

\subsection{Inquiry}\label{sec:inquiry}
Although in Sections \ref{sec:summarization}  we have already enhanced the VLM agent's capabilities using the strong prompt $p_L^{\text{sum}}$ (see supplementary material A.4), due to the complexity of the image information andthe VLM agent’s limitations with long texts, the final extracted information may still be insufficient and even contain conflicting details.

To solve this issue, we prompt the LLM in Section \ref{sec:summarization} to generate a question \(q\), which describes which privacy information is missing and causing the reasoning failure. Note that \(q\) may not always be generated; it will only appear if the information extraction is incomplete. If the extraction is complete, \(q\) will be empty. To resolve \(q\), we need to perform a secondary extraction of the image information. We use the VLM agent to reason by concatenating \(q\) with all the images and following the same approach as in Section \ref{sec:extraction}, we additionally concatenate 
\(q\) to the input, and the input now consists of all images instead of just one. Note that \(q\) focuses on a specific attribute, making the task simpler. For example: ``Which scenes in these photos could determine the person’s gender?'' Even when all the photos are input at once, the task can still be completed successfully.

The question \(q\) can be formulated as:
\begin{equation}
V_{\text{revised}} = \phi_V\left(p^{\text{inquiry}}_ V\oplus q \oplus ({F_i})_{i=1}^{i=N_{\text{origin}}}\right)
\end{equation}
where \(q\) is the generated question, $\phi_V$ is the VLM agent, and \(p_\mathrm{V}^\mathrm{inquiry}\) is the question generation prompt template.

\subsection{Decision Making}
In Section \ref{sec:inquiry}, after obtaining the missing information, we perform the final reasoning based on the previous inferences. The final step of the \sys framework is to summarize the results using the reasoning from Sections \ref{sec:summarization} and \ref{sec:inquiry}, and then determine the final privacy attribute values according to the privacy evaluation standards. 

This can be expressed as:
\begin{equation}
\hat{A}^{(\text{final})} = \phi_L\left(p^{\text{decision}}_L \oplus \left[\hat{A}^{(\text{init})}, V_{\text{revised}}\right]\right)
\end{equation}
where $\hat{A}^{(\text{final})}$ represents the final inferred privacy attribute value, $\phi_L$ is the LLM agent, \(p_\mathrm{L}^\mathrm{decision}\) is the judgment prompt template, and \(V_{\mathrm{revised}}\) is the information obtained from the secondary extraction.

\section{Evaluation}

\subsection{Experimental Setups}
\noindent \textbf{\sys Implementation.} The VLMs we used include four proprietary models: Gemini-2.0-Pro, Llama-3.1-405B, ChatGPT-4o , and Claude-3.7-Sonnet, as well as two open-source models: QwQ-32B and Llama-13B. Additionally, we employed four system prompts (see Appendix E.2) to assist the VLM and LLM agentic framework in performing more effectively.

\medskip \noindent \textbf{Baseline Comparison.} We implemented a series of potential methods based on state-of-the-art VLMs and compared them fairly with our \sys framework. We used advanced prompts to achieve the results, which correspond to the strong prompt in Table~\ref{table:main_table}. The VLMs used include the proprietary models Gemini-2.0-Pro, Llama-3.1-405B [14], ChatGPT-4o [1], and Claude-3.7-Sonnet, as well as open-source models like QwQ-32B and Llama-13B. The direct input for each individual in \data was fed into the VLM agents to predict evaluation metrics. For all these baseline methods, we adopted the same system prompts and hyperparameter settings as the final step in the \sys framework.

\begin{table*}[t!]\centering
\renewcommand{\arraystretch}{0.85} 
\setlength{\tabcolsep}{1.8pt}
\setlength{\abovecaptionskip}{0pt}% 
\setlength{\belowcaptionskip}{0pt}%
\caption{Performance comparison between \sys and other baseline approaches in private attribute profiling.}\label{table:main_table}

\begin{tabular}{llcccccccccccccc}
\toprule
\multicolumn{2}{c}{\multirow{2}{*}{Model}} & \multicolumn{3}{c}{Personal Basic} & \multicolumn{3}{c}{Health Condition} & \multicolumn{3}{c}{Social Attributes} & \multicolumn{3}{c}{Psychological Traits} & \multirow{2}{*}{Average} & \\ 
\cmidrule(lr){3-5} \cmidrule(lr){6-8} \cmidrule(lr){9-11} \cmidrule(lr){12-14}
\multicolumn{2}{c}{} & AG & SE & RE & DR & HS & HH & CA & IN & OC & MBTI & SA & LT & & \\
\midrule
\multirow{3}{*}{Llama-3.1-405B}  

& Strong prompt$^\dagger$ & 81.3 & 83.2 & 79.2 & 77.7 & 76.9 & 82.1 & 72.9 & 76.4 & 78.8 & 55.1 & 75.3 & 76.2 &76.2 & \\
 & \sys(ours) & 88.9 & 89.7 & 87.9 & 85.5 & 86.9 & 88.5 & 90.5 & 90.4 & 90.1 & 65.8 & 88.9 & 84.6 &86.4 & \\
  & $\Delta$ & \cellcolor{red!25} $7.6\uparrow$ & \cellcolor{blue!25} $6.5\uparrow$ & \cellcolor{red!25} $8.7\uparrow$ & \cellcolor{blue!25} $7.8\uparrow$ & \cellcolor{red!25} $8.0\uparrow$& \cellcolor{blue!25} $3.4\uparrow$ &\cellcolor{red!25} $17.6\uparrow$ &\cellcolor{blue!25} $14.0\uparrow$ &\cellcolor{red!25} $11.3\uparrow$ & \cellcolor{blue!25} $10.7\uparrow$ & \cellcolor{red!25} $13.6\uparrow$ & \cellcolor{blue!25} $12.4\uparrow$ &\cellcolor{red!25} $10.2\uparrow$ & \\
\hline
\multirow{3}{*}{ChatGPT-4o}

 & Strong prompt$^\dagger$  & 80.2 & 86.3 & 81.2 & 76.5 & 78.4 & 79.4 & 73.4 & 73.2 & 80.3 & 54.3 & 75.8 & 72.1 &75.9 & \\
 & \sys(ours) & 87.8 & 90.2 & 86.8 & 84.5 & 88.3 & 87.5 & 87.8 & 87.1 & 86.9 & 62.7 & 85.1 & 82.6 &84.7 & \\
  & $\Delta$ & \cellcolor{red!25} $7.6\uparrow$ & \cellcolor{blue!25} $3.9\uparrow$ & \cellcolor{red!25} $5.6\uparrow$ & \cellcolor{blue!25} $8.0\uparrow$ & \cellcolor{red!25} $9.9\uparrow$& \cellcolor{blue!25} $8.1\uparrow$ &\cellcolor{red!25} $14.4\uparrow$ &\cellcolor{blue!25} $13.9\uparrow$ &\cellcolor{red!25} $6.6\uparrow$ & \cellcolor{blue!25} $8.4\uparrow$ & \cellcolor{red!25} $9.3\uparrow$ & \cellcolor{blue!25} $10.5\uparrow$ &\cellcolor{red!25} $9.2\uparrow$ & \\
 \hline
\multirow{3}{*}{Gemini-2.0-pro}

 & Strong prompt$^\dagger$ & 82.9 & 87.9 & 82.9 & 79.7 & 79.1 & 81.3 & 76.1 & 77.9 & 82.1 & 58.2 & 77.8 & 78.1 &78.6 & \\
 & \sys(ours) & 92.4 & 92.3 & 93.2 & 85.7 & 87.3 & 89.7 & 87.7 & 93.3 & 89.1 & 67.7 & 87.9 & 91.6 & 88.1 & \\
  & $\Delta$ & \cellcolor{red!25} $9.5\uparrow$ & \cellcolor{blue!25} $4.4\uparrow$ & \cellcolor{red!25} $10.3\uparrow$ & \cellcolor{blue!25} $6.0\uparrow$ & \cellcolor{red!25} $8.2\uparrow$& \cellcolor{blue!25} $8.4\uparrow$ &\cellcolor{red!25} $11.6\uparrow$ &\cellcolor{blue!25} $15.4\uparrow$ &\cellcolor{red!25} $7.0\uparrow$ & \cellcolor{blue!25} $9.5\uparrow$ & \cellcolor{red!25} $10.1\uparrow$ & \cellcolor{blue!25} $13.5\uparrow$ &\cellcolor{red!25} $9.5\uparrow$ & \\
 \hline
\multirow{3}{*}{Claude-3.7-Sonnet} 

&Strong prompt$^\dagger$  & 85.6 & 90.7 & 86.3 & 77.9 & 81.1 & 82.7 & 74.2 & 79.1 & 81.0 & 59.2 & 79.9 & 78.9 &79.7 & \\
 & \sys(ours) & 96.1 & 91.4 & 96.3 & 87.9 & 87.8 & 93.7 & 88.9 & 93.6 & 94.3 & 70.2 & 89.7 & 96.8 &90.5 & \\
  & $\Delta$ & \cellcolor{red!25} $10.5\uparrow$ & \cellcolor{blue!25} $0.7\uparrow$ & \cellcolor{red!25} $10.0\uparrow$ & \cellcolor{blue!25} $10.0\uparrow$ & \cellcolor{red!25} $6.7\uparrow$& \cellcolor{blue!25} $11.0\uparrow$ &\cellcolor{red!25} $13.3\uparrow$ &\cellcolor{blue!25} $14.5\uparrow$ &\cellcolor{red!25} $13.3\uparrow$ & \cellcolor{blue!25} $11.0\uparrow$ & \cellcolor{red!25} $9.8\uparrow$ & \cellcolor{blue!25} $17.9\uparrow$ &\cellcolor{red!25} $10.8\uparrow$ & \\
\bottomrule
\end{tabular}
\begin{tablenotes}[]
    \item[] \vspace{-2pt}\hspace{-2pt}\small 
       (i) $\dagger$: Strong prompt denotes best inference accuracy among all baselines, using a fine-tuned prompt on a single model. (ii)
       $\Delta$: Absolute accuracy gain of HolmesEye over the Strong prompt (in percentage points).
\end{tablenotes}
\vspace{-10pt}
\end{table*}

\medskip \noindent \textbf{Evaluation Metrics.} We categorized sensitive attributes into four types and used different metrics: 1) Qualitative attributes — SE; 1) Quantitative attributes — HS, CA, IC, SA, LT, DR, HH; 7) Ambiguous attributes — RE, OC, MBTI, AG; 4). We used absolute error accuracy for qualitative attributes to evaluate the experimental results. For quantitative attributes, we measured the relative error accuracy to evaluate the difference between the inferred attribute values and the actual values. The degree of variation for each attribute is detailed in Figure 2 of Chapter 4. To ensure objective evaluation of ambiguous attributes, we used the most powerful LLM, Claude-3.7-Sonnet, to assess ambiguity, generating a similarity score on a five-level scale from 1 to 0, with steps of 0.25. All attribute scores are within the range of 0 to 1, where higher values are better, and we multiplied them by 100.

\subsection{Risks of Private Image Attribute Profiling}
\noindent \textbf{Main Results.} 
Experimental results of using MLLM to analyze sensitive attributes in image data are shown in Table~\ref{table:main_table}. As indicated, the proposed \sys framework consistently outperforms all baseline methods across all attributes for all VLM models. When the same VLM is used, the average inference accuracy improves by up to 10.7\% (absolute error), and when different VLMs are used, the improvement reaches up to 14.5\% (absolute error). Here, the average inference accuracy quantifies the similarity between the inferred curve at the individual level and the true values. These results demonstrate that \sys achieves the most accurate reconstruction of sensitive attribute profiles. Table~\ref{table:main_table} also reveals several key insights. First, MLLM can effectively infer sensitive attributes from image data. However, for relatively abstract social and Psychological traits, the inference performance of MLLM is less than ideal. Even for Claude-3.7-sonnet, which has the highest inference capability, the accuracy for Social Attributes (OC) reaches only 81\%, while for Personal Basic (SE), the highest accuracy is 90.7\%. This indicates that MLLM struggles to infer relatively abstract privacy attributes effectively.
% \begin{table}[]
% \caption{Ablation studies of different phases in the proposed  framework.}
% \begin{tabular}{lllll}

% \hline
% model    & Claude-3.7-sonnet & Chatgpt-4o & Qwq-32b & Llama-13b \\ \hline
% accuracy & 79.7                    & 75.9                     & 76.5                & 71.3                  \\ \hline
% \end{tabular}
% \end{table}

\begin{table}[]
\setlength{\tabcolsep}{5pt}
\setlength{\abovecaptionskip}{0pt}% 
\setlength{\belowcaptionskip}{0pt}%
\caption{\sys can enhance profiling ability of open-source models, on par with SOTA commercial VLM+LLMs.}\label{table:little_model}
\begin{tabular}{llc}
\hline
\textbf{Model} & \textbf{Strategy}& \textbf{Accuracy (\%)} \\
\hline
Claude-3.7-sonnet & Strong prompt & 79.7 \\
ChatGPT-4o & Strong prompt & 75.9 \\
\cellcolor{gray!20}QwQ-32B & \cellcolor{gray!20}Strong prompt & \cellcolor{gray!20}70.3 \\
\cellcolor{gray!20}Llama-13B & \cellcolor{gray!20}Strong prompt& \cellcolor{gray!20}69.1 \\
\cellcolor{red!25}QwQ-32B & \cellcolor{red!25}\sys  & \cellcolor{red!25}78.5 \\
\cellcolor{red!25}Llama-13B & \cellcolor{red!25}\sys  & \cellcolor{red!25}76.3 \\
\bottomrule
\end{tabular}
\vspace{-12pt}
\end{table}

\medskip \noindent \textbf{Impact of Model Parameter Size.} 
To investigate 
the performance of \sys with smaller models, we used open-source models for both the VLM and LLM components. Specifically, we considered two MLLMs --- Llama-13B and QwQ-32B --- for full-attribute profiling within the \sys framework, comparing their performance against the baseline versions of proprietary models. The proprietary models chosen for comparison were the most powerful reasoning models, Claude-3.7-sonnet and the relatively strong ChatGPT-4o. Detailed experimental results are shown in Table~\ref{table:little_model}.

The results indicate that, in general, QwQ-32B outperformed the baseline version of ChatGPT-4o in terms of average accuracy when used within our framework, while also showing mixed results against Claude-3.7-sonnet in two attribute categories. Although the average accuracy of Llama-13B is somewhat lower than Claude-3.7-sonnet, it still surpasses ChatGPT-4o.
From these findings, we conclude that MLLMs with moderate capabilities are sufficient to form an effective \sys framework. Additionally, the \sys framework significantly enhances the image privacy reasoning ability of MLLMs, proving to be both powerful and versatile.

\begin{table*}[h]
\centering
\renewcommand{\arraystretch}{0.85} 
\setlength{\abovecaptionskip}{0pt}% 
\setlength{\belowcaptionskip}{0pt}%
\caption{Ablation studies of different phases in \sys  framework.}\label{table:model_prompt}
\begin{tabular}{l|cccccccccccc|c}
\toprule
\multicolumn{1}{c|}{Model/Attributes}& AG & SE & RE & DR & HS & HH & CA & IN & OC & MBTI & SA & LT &  Average \\ \midrule
\sys w/o Extraction      &93.3 &88.1 &91.5 &85.3 &83.4 &88.1 &85.6 &79.9 &88.9 &62.9 &84.4 &90.4 &85.1     \\  \midrule
\sys w/o Analysis       &93.1 &90.4 &95.0 &85.1 &84.0 &91.0 &80.9 &85.9 &79.6 &61.9 &83.5 &88.1 &84.9     \\  \midrule
\sys w/o Summarization       &91.5 &88.4 &92.1 &86.2 &84.3 &86.9 &86.2 &87.9 &89.2 &66.5 &85.2 &91.3 &86.3     \\ \midrule
\sys w/o Inquiry and Decision      &90.5 &88.5 &89.1 &85.2 &86.0 &88.7 &85.2 &80.9 &86.1 &64.5 &86.2 &89.6 &85.0     \\ \midrule
\rowcolor{gray!20 }  \sys (full)       & 96.1 & 91.4 & 96.3 & 87.9 & 87.8 & 93.7 & 88.9 & 93.6 & 94.3 & 70.2 & 89.7 & 96.8 &90.5     \\ \bottomrule
\end{tabular}

\end{table*}

\medskip \noindent \textbf{Impact of Numbers for Inter-Image Analysis.}\label{exp:num}
To demonstrate the optimal group size for \sys's second phase, we conducted a simple test. Specifically, we randomly selected 50 individuals from a pool of 251 and tested them using the most powerful VLM model, Claude-3.7-sonnet. For this experiment, we set the group sizes to 2, 3, 5, and 7 to determine the optimal number of images for extracting information between the images. The experimental results are shown in Table~\ref{table:model_prompt}. 
From Table~\ref{table:model_prompt}, we observe that the highest accuracy was achieved when the group size was set to 3. By analyzing intermediate files, we found that when the group size was increased to 5 or 7, the performance of Claude-3.7-sonnet decreased due to the larger number of input images. Although a greater number of possible combinations became available, the model’s reasoning ability for each individual combination was diminished.

\medskip \noindent \textbf{Impact of Different Sensitive Attributes.}
In Section \ref{sec:analysis}, privacy attributes are divided into four categories: personal information, health status, psychological characteristics, and social characteristics. Among them, personal information and health status can usually be directly observed from images, making it easier for privacy detection systems to identify them in images. In contrast, psychological characteristics and social characteristics often rely on contextual or inferred information, making them more difficult to detect. As shown in Table 1, when using our proposed framework or baseline methods, the accuracy of detecting psychological and social attributes is consistently lower than the accuracy of detecting personal information and health status attributes. This indicates that since privacy features related to personal information and health status are relatively easier to express in images, while psychological and social attributes are not directly displayed, detection models find it relatively difficult to process these attributes.

\medskip \noindent \textbf{Impact of Total Image Numbers.} As outlined in section \ref{sec:threat_model}, each individual within the dataset is associated with multiple images. For the purpose of attack scenarios, it is naturally advantageous to have a larger number of images. However, in order to assess the ability of VLMs to infer privacy-related information from groups of images, it is necessary to control the number of images used, while still ensuring that all relevant attributes can be accurately inferred. To achieve this balance, we began by selecting a set of photos provided by 50 volunteers. Various numbers of photos were then utilized, and baseline methods were applied to infer privacy information from these images.

The results presented in Table 5 demonstrate that when the number of photos was limited, the inference accuracy was significantly reduced due to a considerable amount of missing data. This highlights the challenge of accurately inferring privacy information with insufficient data. However, when 10 photos were employed, a reasonable balance was achieved. This number of photos allowed for a satisfactory inference accuracy, while simultaneously minimizing the number of images, thus aligning with the goal of dataset construction.
This outcome supports the rationale underlying the dataset design, as it demonstrates that selecting 10 photos per individual optimizes the trade-off between inference performance and data minimization. This approach ensures that the VLM can make effective privacy inferences without relying on an excessive number of images, which would be more resource-intensive and potentially lead to unnecessary privacy concerns.
% \begin{table}[]
% \centering
% \setlength{\tabcolsep}{3.0pt}
% \setlength{\abovecaptionskip}{0pt}% 
% \setlength{\belowcaptionskip}{0pt}%
% \caption{The impact of the number of \data single-person images on the results.}\label{table:figure}
% \begin{tabular}{lccccc}
% \toprule
% Model              & 5 imgs   & 7 imgs   & 9 imgs    & \textbf{10 imgs} &11 imgs   \\
% \midrule
% Claude-3.7-sonnet & 62.2 & 72.5 & 76.2 & \textbf{79.7} &80.1  \\
% ChatGPT-4o        & 51.2 & 64.7 & 72.3 & \textbf{75.9} &76.2\\
% Llama-3.1         & 54.5 & 66.4 & 74.4 & \textbf{76.2} &76.6\\
% Gemini-2.0-pro    & 58.7 & 68.1 & 76.1 & \textbf{78.6} &79.1\\
% \bottomrule
% \end{tabular}
% \end{table}

\begin{table}[]
\centering
\setlength{\tabcolsep}{3.0pt}
\setlength{\abovecaptionskip}{0pt}% 
\setlength{\belowcaptionskip}{0pt}%
\caption{The impact of the number of \data single-person images on the results.}\label{table:figure}
\begin{tabular}{lccccc}
\toprule
Model              & 5 imgs   & 7 imgs   & 9 imgs    & \cellcolor{gray!20}10 imgs&11 imgs   \\
\midrule
Claude-3.7-sonnet & 62.2 & 72.5 & 76.2 & \cellcolor{gray!20} \textbf{79.7} & 80.1  \\
ChatGPT-4o        & 51.2 & 64.7 & 72.3 & \cellcolor{gray!20} \textbf{75.9} &76.2\\
Llama-3.1-405B         & 54.5 & 66.4 & 74.4 & \cellcolor{gray!20} \textbf{76.2} &76.6\\
Gemini-2.0-pro    & 58.7 & 68.1 & 76.1 & \cellcolor{gray!20} \textbf{78.6} &79.1\\
\bottomrule
\end{tabular}
\vspace{-10pt}
\end{table}

\noindent \textbf{Ablation Study.}  To validate the effectiveness of each stage in the \sys framework, we conducted a comprehensive ablation study. To ensure that the \sys framework remains effective in the future, we selected the most powerful reasoning model, Claude-3.7-sonnet, for the ablation process.
(1) We removed the extraction component from the ALM prompt. The corresponding experimental results are shown in the “\sys w/o Extract” cell in Table \ref{table:model_prompt}. The results indicate that removing extraction from the ALM prompt significantly decreases reasoning accuracy. This finding confirms the importance of using LLM-generated prompts for extraction during the reasoning process.
(2) We disabled the Analysis phase, meaning that the system no longer groups or extracts information from the image. A comparison of the results between “\sys w/o Analysis” and “\sys (full)” reveals a significant performance drop in inferring certain attributes, particularly social and Psychological traits. This observation highlights the necessity of the Forensics phase for accurate attribute inference and underscores the importance of the Analysis phase for reasoning about more abstract attributes. 
(3) We disabled the Summary phase, allowing the initial reasoning results from the VLM to directly input into the final output generated by the LLM. The continued decline in accuracy across all attributes emphasizes the crucial role of the Summary phase in ensuring accurate attribute inference by the LLM.  
(4) We removed both the Inquiry and Judgment phases. Since these two phases are deeply interconnected, omitting either results in performance similar to a direct summary. Therefore, we performed their combined ablation. Comparing the results of ``\sys w/o Inquiry-Judgment'' and ``\sys (full)'' shows a significant drop in performance, indicating that the Inquiry-Judgment phases are essential for achieving optimal results.

\begin{figure}[t]
  \centering
  \includegraphics[width=0.9\columnwidth]{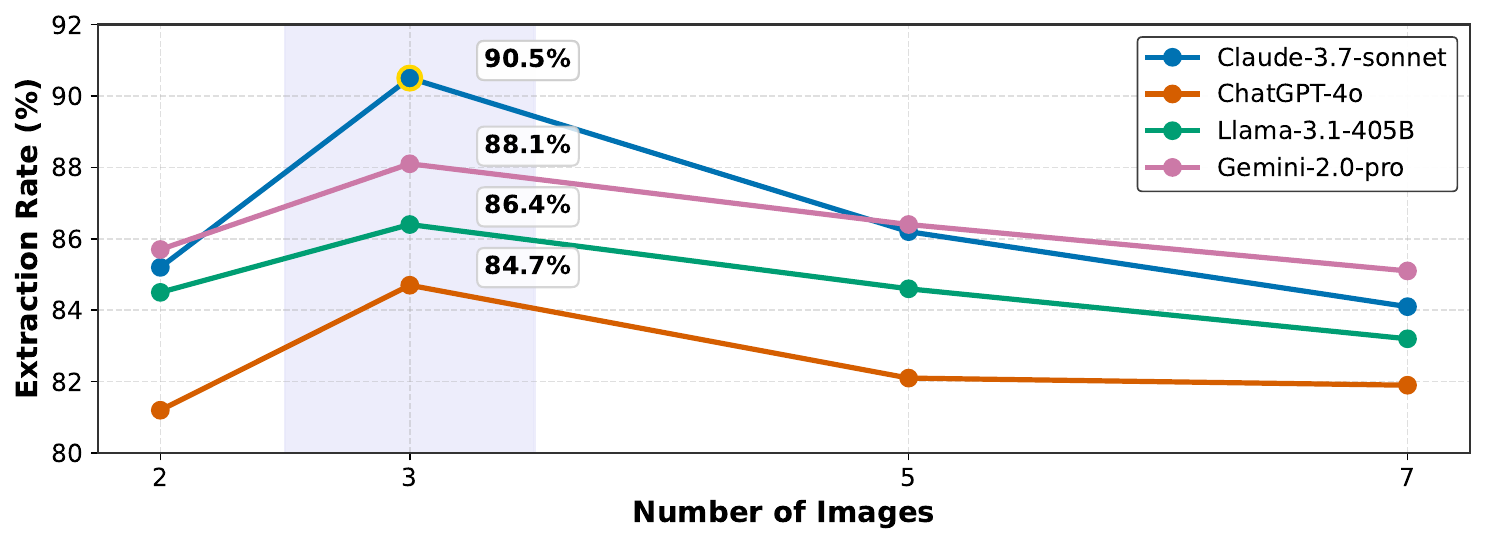}
  \vspace{-10pt}
  \caption{The impact of the number of PAPI single-person
images on the results.}
  \vspace{-20pt}
  \label{fig:Impact of image numbers.}
\end{figure}

\subsection{Human Evaluation}

\noindent \textbf{Settings.} To further evaluate the effectiveness and efficiency of using MLLMs to extract sensitive attributes from images, we conducted a human evaluation in accordance with IRB regulations and privacy protection policies. In this study, we recruited 50 participants with higher education backgrounds. Additionally, most participants were familiar with and accustomed to using various search engines and services. For this study, we randomly selected 50 simulated individuals from the PAPI dataset. Participants were asked to observe ten images belonging to each individual and infer privacy-related attributes. They were allowed to review the images repeatedly as needed and were permitted to use search engines or retrieve relevant information to make more accurate inferences.

\medskip \noindent \textbf{Comparison between Humans and MLLM Agents.} Table \ref{table:human} presents experimental results comparing real humans with high-performance MLLM agents in analyzing sensitive individual attributes. Notably, our \sys framework demonstrates dual superiority: it not only achieves the highest average inference accuracy of 90.5\%—surpassing humans by 15.0\% points and the next-best model by 5.8\% points—but also completes inference in just 3 minutes and 23 seconds. This represents a \textbf{3.45-fold speedup over human performance}, showcasing unprecedented efficiency in sensitive attribute analysis. While ChatGPT-4o shows marginally faster time, its accuracy lags behind our framework across all evaluated dimensions. These findings underscore the critical risk posed by MLLMs: their combination of superhuman accuracy and dramatically reduced inference time creates significant privacy vulnerabilities in automated sensitive attribute analysis from images.

% \begin{table}
% \centering
% \begin{tabular}{lccccc}
% \toprule
% Model & Personal Basic & Health Status & Social Attributes & Psychological Traits & Avg \\
% \midrule
% human & 78.3 & 78.9 & 72.2 & 75.5 & 76.5 \\
% Claude-3.7-sonnet & 94.6 & 89.8 & 92.3 & 85.6 & 92.2 \\
% Chatgpt-4o & 88.3 & 86.8 & 87.3 & 76.8 & 87.5 \\
% \bottomrule
% \end{tabular}
% \end{table}
\begin{table}
\centering
\renewcommand{\arraystretch}{0.85} \setlength{\tabcolsep}{2.0pt}
\setlength{\abovecaptionskip}{0pt}% 
\setlength{\belowcaptionskip}{0pt}%
\caption{Comparison between \sys framework and real humans in inference accuracy.}\label{table:human}
\begin{tabular}{lccc}
\toprule
Evaluation & Human & \cellcolor{gray!20}Claude-3.7-Sonnet & ChatGPT-4o \\
\midrule
Personal Basic & 86.3 & \cellcolor{gray!20}\textbf{94.6} & 88.3 \\
Health Status & 78.9 & \cellcolor{gray!20}\textbf{89.8} & 86.8 \\
Social Attributes & 69.3 & \cellcolor{gray!20}\textbf{92.3} & 87.3 \\
Psychological Traits & 67.5 & \cellcolor{gray!20}\textbf{85.6} & 76.8 \\
\midrule
Average(\%) & 75.5 & \cellcolor{gray!20}\textbf{90.5} & 84.7 \\
Time(mm'ss'') & 11'29'' &\cellcolor{gray!20}\textbf{3'23''} &3'17''\\
\bottomrule
\end{tabular}
\vspace{-10pt}
\end{table}

\subsection{Comparison with Traditional CV Technique}

Traditional computer vision (CV) models have proven satisfactory in many cases, particularly when labeling the sex and age of a given facial image. To evaluate whether the newer MLLM agentic framework provides substantial improvements, we compare \sys against a state-of-the-art age inference model, Deep Age Attributes (DAA)~\cite{CVPR-DAA}. We do not compare other attributes like MBTI, because traditional rely on direct facial images, while the majority of the data in the \data dataset consists of indirect images, making age profiling a more suitable and fair comparison. In this experiment, we use the simplest implementation of \sys, which is based on a VLM agent with a naive prompting strategy, analyzing ten images simultaneously. We then compare its performance to that of the DAA model.

As shown in Table~\ref{tab:CV_class_model}, our VLM-based approach outperforms the specialized DAA model by 13.8\% in accuracy, despite using a much simpler setup. This result demonstrates that, even with minimal prompt engineering, our method can offer a significant advantage in age inference accuracy. There are three key reasons for this improvement. 1) Unlike traditional models, \sys does not require extensive training or fine-tuning. This makes it more adaptable across various attributes and contexts. 2) Even for the specific task (age inference) that DAA is designed for, our VLM-based method outperforms it. This suggests that the potential benefits of \sys are even more pronounced for more abstract or complex attributes. 3) Traditional models, like DAA, are limited to explicit visual cues, such as facial features. However, \sys can infer attributes from indirect contextual elements in images, such as environmental factors or clothing styles, that might not be directly related to the attribute being inferred but still contribute to the accuracy of the result.

\begin{table}[t]
    \renewcommand{\arraystretch}{0.85} 
    \setlength{\abovecaptionskip}{0pt}% 
    \setlength{\belowcaptionskip}{0pt}%
    \caption{Comparison of age inference accuracy between the naive VLM-based approach and the traditional DAA model.}\label{tab:CV_class_model}
    \centering
    \begin{tabular}{ccc}
        \toprule
        Model & \cellcolor{gray!20}VLM-only (Naive prompt) & DAA \\
        \midrule
        Accuracy (\%) & \cellcolor{gray!20}\textbf{70.2} & 56.4 \\
        \bottomrule
    \end{tabular}
    \vspace{-15pt}
\end{table}

\section{Discussion}
\noindent \textbf{Limitations.} Despite \sys's strong performance in profiling both concrete and abstract attributes from personal image collections, several limitations persist. The framework's inference accuracy for abstract traits—such as MBTI types and leadership tendencies—remains consistently lower than for visually grounded attributes like age or health status. This disparity likely stems from the scarcity of explicit visual cues for abstract traits, coupled with the model's tendency to generate conservative predictions when faced with ambiguous evidence. Such observations suggest current MLLMs struggle with reliable individualized inferences in contexts requiring sophisticated psychological understanding. Our analysis also revealed systemic biases, with models like Claude-3.7-Sonnet and ChatGPT-4o disproportionately classifying individuals into higher socioeconomic categories.

\medskip \noindent \textbf{Potential Countermeasures.} Safety calibration of VLMs warrants high priority, as our research demonstrates their capability to infer sensitive attributes from visual data with concerning accuracy. However, implementing effective safety measures faces some challenges. First, supervised fine-tuning and RLHF techniques require large-scale, high-quality datasets to comprehensively control model outputs—datasets that are particularly difficult to collect for privacy-related image content. Second, safety calibration may compromise general functionality since core VLM tasks like image description inherently rely on features that overlap with privacy-relevant information. Despite these challenges, safety alignment remains achievable~\cite{liuyue_GuardReasoner}. Future approaches could leverage differential privacy, adversarial training, and feature decoupling during training or fine-tuning to reduce models' dependence on privacy-sensitive features while preserving their core functionality.
\section{Conclusion}
In this work, we conducted a comprehensive study on image privacy vulnerabilities introduced by MLLMs. We addressed the lack of suitable evaluation resources by creating PAPI, a benchmark dataset specifically designed for analyzing private attribute inference from personal image collections. We then developed HolmesEye, a framework that synergistically combines vision-language models and large language models to enhance privacy attribute inference capabilities. Our extensive experiments demonstrate that HolmesEye significantly outperforms baseline methods in extracting sensitive attributes from image sets, with particularly concerning implications for abstract attributes that users may not realize are discernible from their photos. These findings highlight the urgent need for privacy-preserving mechanisms specifically designed to counter sophisticated attribute inference attacks from visual data.

\bibliographystyle{ACM-Reference-Format}
% \balance
\bibliography{reference/reference.bib}

\end{document}